\def\year{2022}\relax
\documentclass[letterpaper]{article} % DO NOT CHANGE THIS
\usepackage{aaai22}  % DO NOT CHANGE THIS
\usepackage{times}  % DO NOT CHANGE THIS
\usepackage{helvet}  % DO NOT CHANGE THIS
\usepackage{courier}  % DO NOT CHANGE THIS
\usepackage[hyphens]{url}  % DO NOT CHANGE THIS
\usepackage{graphicx} % DO NOT CHANGE THIS
\usepackage{natbib}  % DO NOT CHANGE THIS AND DO NOT ADD ANY OPTIONS TO IT
\usepackage{caption} % DO NOT CHANGE THIS AND DO NOT ADD ANY OPTIONS TO IT
\DeclareCaptionStyle{ruled}{labelfont=normalfont,labelsep=colon,strut=off} % DO NOT CHANGE THIS
\usepackage{algorithm}
\usepackage{algorithmic}
\usepackage{multirow}
\usepackage{booktabs}
\usepackage{hyperref}

\usepackage{newfloat}
\usepackage{listings}
\usepackage{amssymb}

\floatstyle{ruled}
\newfloat{listing}{tb}{lst}{}
\floatname{listing}{Listing}
\title{Fully Attentional Network for Semantic Segmentation}
\author{
    Qi Song,\textsuperscript{\rm 1,}\textsuperscript{\rm 2}
    Jie Li,\textsuperscript{\rm 1,}\textsuperscript{\rm 2}$^\ast$  Chenghong Li,\textsuperscript{\rm 1}\thanks{These authors contributed equally.~}
    Hao Guo,\textsuperscript{\rm 1,}\textsuperscript{\rm 2}
    Rui Huang\textsuperscript{\rm 1}\thanks{Corresponding author.}
}
\affiliations{
     \textsuperscript{\rm 1}The Chinese University of Hong Kong, Shenzhen\\
     \textsuperscript{\rm 2}Shenzhen Institute of Artificial Intelligence and Robotics for Society\\
       \{qisong, jieli1, chenghongli, haoguo\}@link.cuhk.edu.cn, 
   ruihuang@cuhk.edu.cn
}

\usepackage{bibentry}
\begin{document}

\maketitle

\begin{abstract}
Recent non-local self-attention methods have proven to be effective in capturing long-range dependencies for semantic segmentation. These methods usually form a similarity map of \(\mathbb{R}^{C\times C}\) (by compressing spatial dimensions) or \(\mathbb{R}^{HW\times HW}\) (by compressing channels) to describe the feature relations along either channel or spatial dimensions, where \(C\) is the number of channels, \(H\) and \(W\) are the spatial dimensions of the input feature map. However, such practices tend to condense feature dependencies along the other dimensions, hence causing attention missing, which might lead to inferior results for small/thin categories or inconsistent segmentation inside large objects. To address this problem, we propose a new approach, namely Fully Attentional Network (FLANet), to encode both spatial and channel attentions in a single similarity map while maintaining high computational efficiency. Specifically, for each channel map, our FLANet can harvest feature responses from all other channel maps, and the associated spatial positions as well, through a novel fully attentional module. Our new method has achieved state-of-the-art performance on three challenging semantic segmentation datasets, i.e., 83.6\%, 46.99\%, and 88.5\% on the Cityscapes test set, the ADE20K validation set, and the PASCAL VOC test set, respectively. Our code will be available at \href{https://github.com/Ilareina/FullyAttentional}{https://github.com/Ilareina/FullyAttentional}.
\end{abstract}

\begin{figure}[t]
\centering
\includegraphics[width=1\columnwidth]{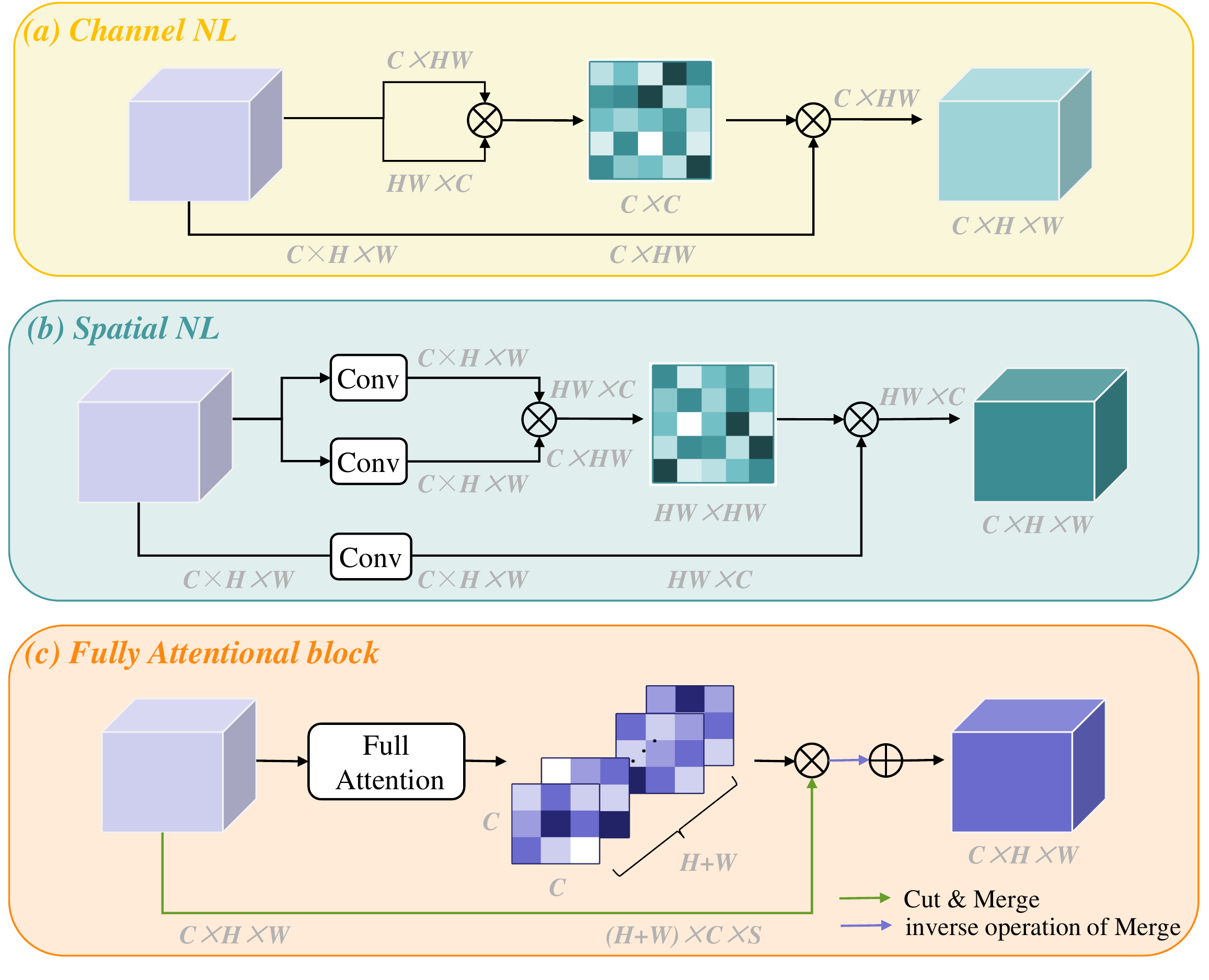} 
\caption{Architectures of Non-Local blocks (NL) and our proposed Fully Attentional block (FLA). The traditional NLs compute a similarity map along channel or spatial dimension, while our FLA generates a similarity map in all dimensions and enables full attentions in the attention map.}
\label{fig1}
\end{figure}

\begin{figure*}[t]
\centering
\includegraphics[width=1\textwidth]{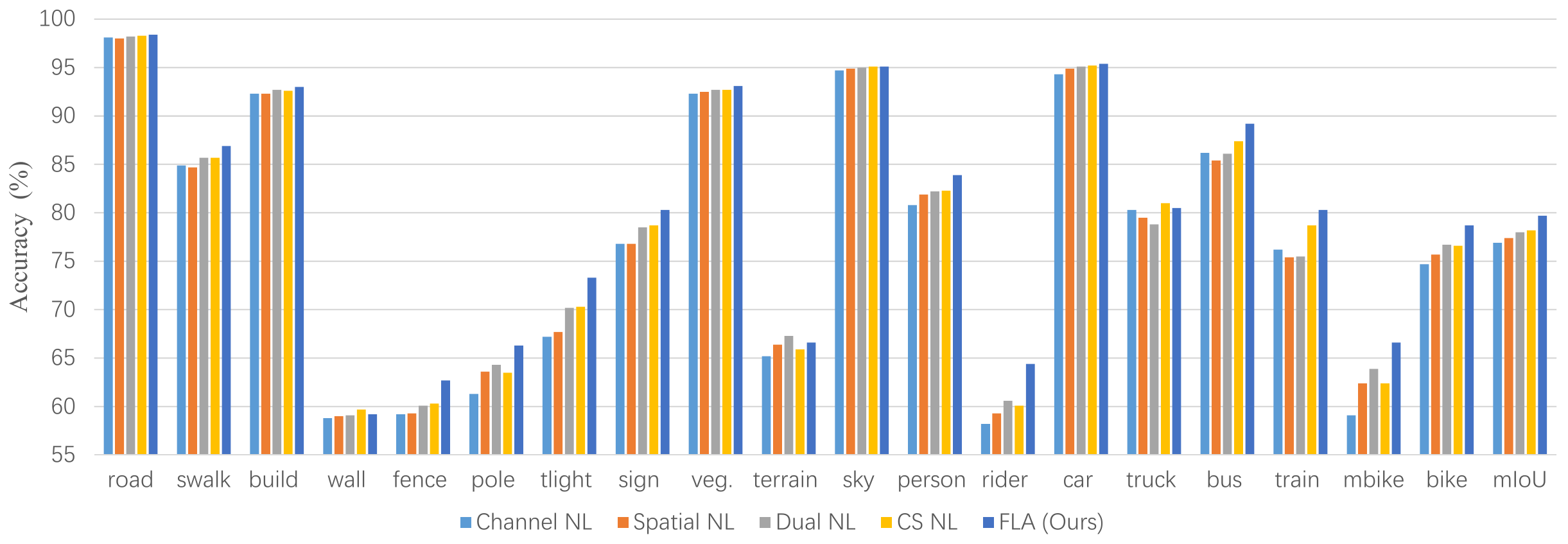} 
\caption{The motivation of our approach and the quantitative evidence for the attention missing issue on the Cityscapes validation set. This observation proves that 1) Spatial NL can enhance the discrimination of details, while Channel NL benefits in maintaining semantic consistency inside large objects, 2) stacking two NL blocks in the network still suffers from the attention missing issue, and 3) our proposed FLA can successfully tackle the attention missing issue since we can achieve better accuracy than that of a single NL block in all classes by modeling full attentions.}
\label{fig2}
\end{figure*}

\section{Introduction}

\noindent Recently, semantic segmentation models achieve great progress by capturing long-range dependencies \cite{zhao2017pyramid,Yang2018DenseASPPFS,Yuan2020ObjectContextualRF,Sun2019DeepHR}, in which Non-Local (NL) based methods are the mainstream \cite{zhao2018psanet, fu2019dual, zhang2019acfnet,Zhu2019AsymmetricNN,Ramachandran2019StandAloneSI}. To generate dense and well-rounded contextual information, NL based models utilize a self-attention mechanism to explore the interdependencies along the channel \cite{Cao2019GCNetNN, zhao2018psanet} or spatial \cite{huang2019ccnet, Yin2020DisentangledNN,Song2021AttaNetAN} dimensions. We denote these two variants of NL block as “Channel NL” and “Spatial NL”, respectively, and the architectures of these two variants are illustrated in Fig.\ref{fig1} (a) and (b). Although these explorations have made impressive contributions to semantic segmentation, one acute issue, i.e., \textbf{attention missing}, was mostly ignored. Take Channel NL for example, the channel attention map \(\mathbb{R}^{C\times C}\)  is generated by the matrix multiplication of two inputs with a dimension of \(C\times HW\) and \(HW\times C\). It can be found that each channel can be connected with all other channel maps while the spatial information will be integrated and each spatial position fails to perceive feature response from other positions during the matrix multiplication. Similarly, interactions among channel dimensions are also missing in the Spatial NL. 

We argue that the attention missing issue would damage the integrity of 3D context information (\(CHW\)) and thus both NL variants can only benefit partially in a complementary way. To verify this hypothesis, we present the per-class comparison results on the Cityscapes validation set in Fig.\ref{fig2}. As shown in the figure, Channel NL gets better segmentation results among large objects, such as \textit{truck}, \textit{bus} and \textit{train}, while Spatial NL performs much better on small/thin categories, e.g., \textit{poles}, \textit{rider} and \textit{mbike}. They both lose precision in some categories due to the mentioned attention missing issue. Besides, we are also curious about whether this issue can be solved by stacking the two blocks sequentially. We denote the parallel connection mode in DANet \cite{fu2019dual} and the sequential Channel-Spatial NL as “Dual NL” and “CS NL”, respectively\footnote{Since there are no convolutional layers in the Channel NL, if we employ Spatial NL before Channel NL (SC NL), the feature weights tend to be either extremely large or extremely small after two consecutive enhancements and the training loss does not converge. So the performance of this connection mode is not reported.}. Intuitively, when two NLs are employed at the same time, the accuracy gain of each class should be no less than that of a single NL. However, it is observed that the performance of Dual NL drops a lot in large objects such as \textit{truck} and \textit{train}, and CS NL gets poor IoU results in some thin categories like \textit{pole} and \textit{mbike}. We can find that both Dual NL and CS NL can only preserve partial benefits brought by either Channel NL or Spatial NL. Therefore, we can conclude that: the attention missing issue hurts the feature representation ability and it cannot be solved by simply stacking different NL blocks. 

Motivated by this, we propose a novel non-local block namely Fully Attentional block (FLA) to efficiently retain attentions in all dimensions. And the workflow is shown in Fig.\ref{fig1} (c). The basic idea is to utilize the global context information to receive spatial responses when computing the channel attention map, which enables full attentions in a single attention unit with high computational efficiency. Specifically, we first enable each spatial position to harvest feature responses from the global contexts with the same horizontal and vertical coordinates. Second, we use the self-attention mechanism to capture the fully attentional similarities between any two channel maps and the associated spatial positions. Finally, the generated fully attentional similarities are used to re-weight each channel map by integrating features among all channel maps and associated global clues. 
 
It should be noted that our method is more effective and efficient than previous works \cite{fu2019dual, Babiloni2020TESATE} when modeling interdependencies in all dimensions. Since we encode spatial interactions into the traditional Channel NL and capture full attentions in a single attention map, our FLA is with high computational efficiency. Specifically, our FLA significantly reduces FLOPs by about \textbf{83\%} and only requires \textbf{34\%} GPU memory usage of DANet in computing both spatial and channel dependencies.

We have carried out extensive experiments on three challenging semantic segmentation datasets and our approach achieves state-of-the-art performance on these experiments. Moreover, our model outperforms other non-local based methods by a large margin with the same backbone network. Our contributions mainly lie in three aspects:
\begin{itemize}
\item Through the theoretical and experimental analysis, we find out the attention missing issue existing in the non-local self-attention methods, which would hurt the integrity of feature representation.
\item We reformulate the self-attention mechanism into a fully attentional manner to generate dense and well-rounded feature dependencies, which addresses the attention missing issue effectively and efficiently. To the best of our knowledge, this paper is the first to achieve full attentions in a single non-local block.
\item We conducted extensive experiments on three challenging semantic segmentation datasets, including Cityscapes, ADE20K, and PASCAL VOC, which demonstrate the superiority of our approach over other state-of-the-art methods.
\end{itemize}

\begin{figure}[t]
\centering
\includegraphics[width=1\columnwidth]{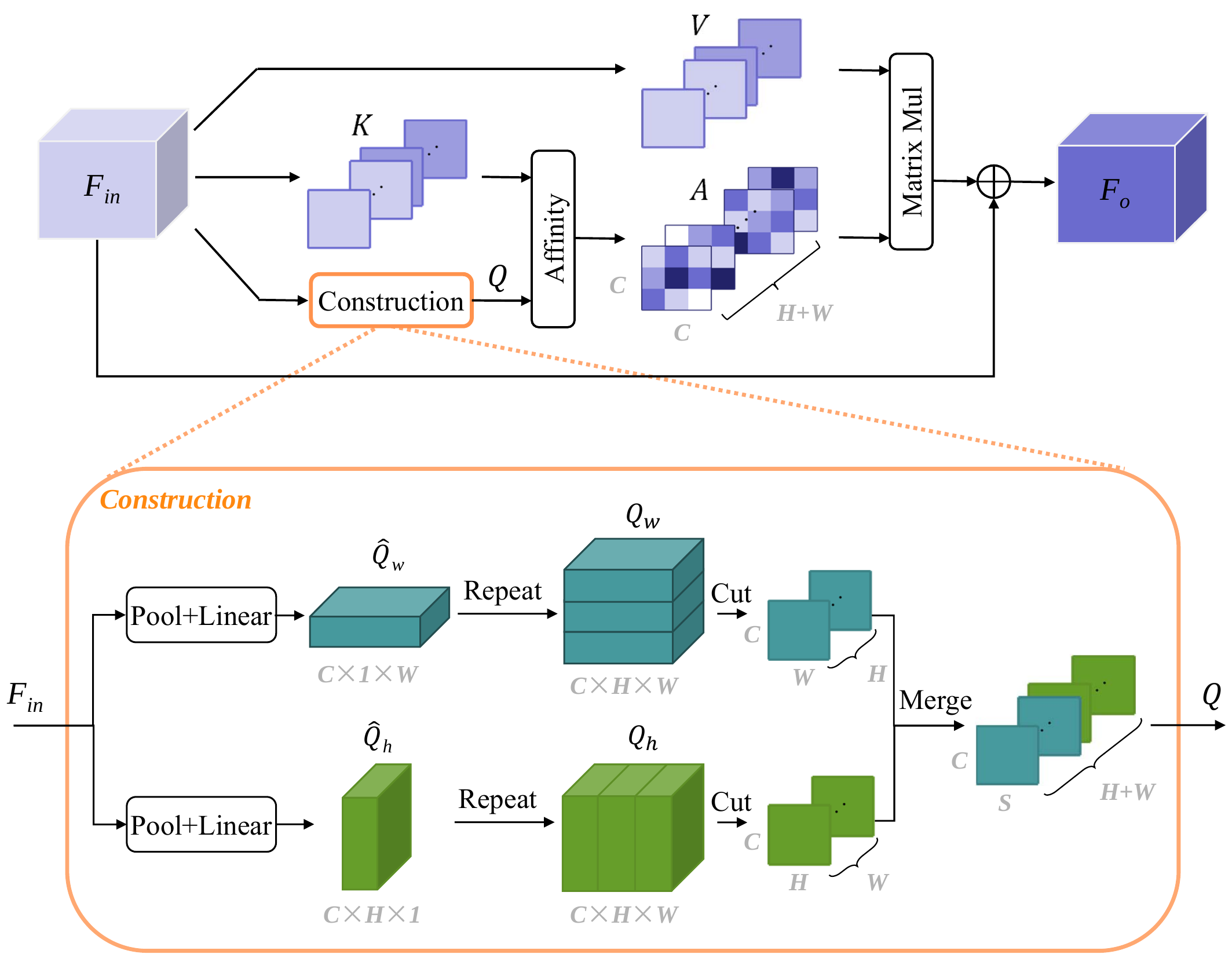} 
\caption{ The details of Fully Attentional block. Since \(H\) equals \(W\) in our implementation, we use the letter \(S\) to represent the dimension after merge for a clear illustration.}
\label{fig3}
\end{figure}

\section{Related Work}
\subsubsection{Semantic Segmentation.}
Semantic segmentation is a vital task in computer vision, which predicts correct semantic labels for all pixels in an image. The traditional classification network based on CNNs can only identify the class of the whole image, not the label of each pixel. Instead of the fully connected layer in the CNN, the FCN \cite{long2015fully} utilizes a convolutional layer to get the segmentation result. UNet \cite{ronneberger2015u} adopts an encoder-decoder structure to recover the detailed information damaged by the step-by-step downsampling operations. To model interdependencies between different channel maps, SENet \cite{hu2018squeeze} produces an embedding of the global distribution of channel-wise feature responses. To enhance the global connections between spatial positions, self-attention based methods are thus proposed to weigh the importance of each spatial position whilst sacrificing the channel-wise attention. Different from these approaches, we argue that the attention missing issue might lead to inconsistent segmentation inside large objects or inferior results for small categories in the semantic segmentation task. Thus in this paper, we consider both channel and spatial dependencies are of equal importance and try to capture both of them in a single attention unit.

\subsubsection{Self-Attention Mechanism. }
Self-attention is initially used for machine translation \cite{Chorowski2015AttentionBasedMF, vaswani2017attention} to capture long-range features. After that, self-attention modules are widely applied in the semantic segmentation field, in which the Non-local network \cite{wang2018non} is the pioneering work. CCNet \cite{huang2019ccnet} harvests the contextual information for each pixel on the criss-cross path. AttaNet \cite{Song2021AttaNetAN} utilizes a striping operation to encode the global context in the vertical direction and then harvests long-range relations along the horizontal axis. OCNet \cite{Yuan2021OCNetOC} utilizes the interlaced self-attention scheme to model both global and local relations. However, these approaches construct a similarity map to leverage relationships along a single dimension, where the dependency along other dimensions is discarded during the matrix multiplication. To generate both spatial-wise and channel-wise attentions, many studies were proposed. DANet \cite{fu2019dual} proposes the position attention module and channel attention module to model dependencies along the spatial and channel dimension respectively. TESA \cite{Babiloni2020TESATE} views the input tensor as a combination of its three-mode matricizations and then captures similarities for each dimension. Although these methods capture relations in all dimensions, they consider different dimensions separately and the attention missing issue still exists in each attention map. To mitigate this issue, we propose a Fully Attentional block to encode both spatial and channel attentions in a single similarity map with high computational efficiency.

\section{Method}

\subsection{Network Architecture}
In this paper, we employ ResNet101 \cite{he2016identity} and HRNetW48 \cite{Sun2019DeepHR} as the backbone network. For ResNet101, dilation convolutions were applied in the last two layers to obtain more detailed information, and the output feature map was enlarged to 1/8 of the input image. Initially, the input image is processed by the backbone network to produce feature maps \(X\). After that, we first apply two convolution layers on \(X\) to reduce the channel dimension and obtain the feature maps \(F_{in}\). Then, the feature maps \(F_{in}\) would be fed into the Fully Attentional block (FLA) and generate new feature maps \(F_{o}\) which aggregate non-local contextual information in all dimensions. Finally, the dense contextual feature \(F_{o}\) is sent to the prediction layer to generate the final segmentation map.

\subsection{Fully Attentional Block}
Previous works try to generate full attentions by applying the attention operation in each dimension in turn, which yields high computational complexity and the single-dimensional attention still overlooks correlations along other dimensions. To capture full attentions in a single attention map with high computational efficiency, we propose a novel non-local block named Fully Attentional block. Specifically, to avoid adding extra computation burden, we try to introduce spatial interactions into channel NL mechanism by utilizing the global average pooling result as the global contextual prior. 

The pipeline of our method is shown in Fig.\ref{fig3}. Given an input feature map \(F_{in}\in \mathbb{R}^{C\times H \times W}\), where \(C\) is the number of channels, \(H\) and \(W\) are the spatial dimensions of the input tensor. First, we feed \(F_{in}\) into two parallel pathways at the bottom (i.e., the \textbf{Construction}), each of which contains a global average pooling layer followed by a Linear layer. When choosing the size of pooling windows, we considered the following two aspects. Firstly, to obtain richer global contextual priors, we choose to use unequal global pooling size in height and width directions rather than kernel windows like \(3\times 3\). Secondly, to make sure that each spatial position is connected with the corresponding global prior with the same horizontal or vertical coordinate, i.e., maintain the spatial consistency when computing channel relations, we choose to keep the length of one dimension constant. Therefore, we employ pooling windows of size \(H\times 1\) and \(1\times W\) in these two pathways respectively. This gives \(\hat{Q}_{w}\in \mathbb{R}^{C\times 1 \times W}\) and \(\hat{Q}_{h}\in \mathbb{R}^{C\times H \times 1}\). After that, we repeat \(\hat{Q}_{w}\) and \(\hat{Q}_{h}\) to form global features \(Q_{w}\in \mathbb{R}^{C\times H \times W}\) and \(Q_{h}\in \mathbb{R}^{C\times H \times W}\). Note that \(Q_{w}\) and \(Q_{h}\) represent the global priors in the horizontal and vertical directions respectively and they will be used to achieve spatial interactions in the corresponding dimension. Furthermore, we cut \(Q_{w}\) along the \(H\) dimension, from which we can generate a group of \(H\) slices with a size of \(\mathbb{R}^{C\times W}\). Similarly, we cut \(Q_{h}\) along the \(W\) dimension. We then merge these two groups to form the final global contexts \(Q\in \mathbb{R}^{(H+W)\times C\times S}\). The cut and merge operations are detailly illustrated in Fig.3.

Meanwhile, we cut the input feature \(F_{in}\) along the \(H\) dimension, yielding a group of \(H\) slices with the size of \(\mathbb{R}^{C\times W}\). Similarly, we do this along the \(W\) dimension. Like the merge process of \(Q\), these two groups are integrated to form the features \(K\in \mathbb{R}^{(H+W)\times S\times C}\). In the same way, we can generate the feature maps \(V\in \mathbb{R}^{(H+W)\times C\times S}\).

After that, we can make each spatial position to receive the feature responses from the global priors in the same row and the same column, i.e., capturing the full attentions \(A\in \mathbb{R}^{(H+W)\times C\times C}\), via the Affinity operation. The Affinity operation is defined as follows: 
\begin{equation}
 A_{i,j}=\frac{exp(Q_{i}\cdot K_{j})}{\sum_{i=1}^C exp(Q_{i}\cdot K_{j})}\,,
\end{equation}
where \(A_{i,j}\in A\) denotes the degree of correlation between the \(i^{th}\) and \(j^{th}\) channel at a specific spatial position.

Then we perform a matrix multiplication between \(A\) and \(V\) to update each channel map with the generated full attentions. After that, we reshape the result into two groups, and each group is with a size of \(\mathbb{R}^{C\times H \times W}\) (i.e., the inverse operation of merge). We sum these two groups to form the long-range contextual information. Finally, we multiply the contextual information by a scale parameter \(\gamma\) and perform an element-wise sum operation with the input feature map \(F_{in}\) to obtain the final output \(F_{o}\in \mathbb{R}^{C\times H \times W}\) as follows:
\begin{equation}
F_{o\,_j}=\gamma \sum_{i=1}^C A_{i,j}\cdot V_{j} + F_{in\,_j}\,,
\end{equation}
where \(F_{o\,_j}\) is a feature vector in the output feature map \(F_{o}\) at the \(j^{th}\) channel map. 

It is noted that different from the traditional Channel NL method which explores only channel correlations by multiplying the spatial information from the same position, our FLA enables spatial connections between different spatial positions, i.e., we exploit full attentions along both spatial and channel dimensions with a single attention map. In this way, our FLA has a more holistic contextual view and is more robust to different scenarios. Moreover, the constructed prior representation brings a global receptive field and helps to boost the feature discrimination ability.

\subsection{Complexity Analysis}

Given a feature map with a size of \(C\times H \times W\), the typical Spatial NL has a computational complexity of \(\textit{O}((HW)^2C)\), and the Channel NL has a computational complexity of \(\textit{O}(C^2HW)\). Both of them can only capture similarities along a single dimension. To model feature dependencies in all dimensions, previous work like DANet applies both Spatial NL and Channel NL to calculate spatial and channel relations separately, which yields higher computational complexity and occupies much more GPU memory. Different from previous works, we achieve full attention in a single NL block and in a more efficient way. Specifically, we utilize the newly constructed global representations to achieve interactions between different spatial positions and collect contextual similarities from all dimensions. And the complexity of our FLA block (both in time and space) is \(\textit{O}(C^2(H+W)S)\). Since \(S=H=W\) in our paper\footnote{We used the square inputs simply for a clear illustration of our method. However, there are no requirements for square inputs in our FLA block.}, our complexity is of the same order with Channel NL and only differs by a small constant.

\section{Experiments}

\begin{table}
\centering
\small
\begin{tabular}{lcc}
\toprule[2pt]
Method & Backbone & mIoU \\
\midrule[2pt]
\textit{Simple Backbone}&&\\
AAF \cite{Ke2018AdaptiveAF}\(^\dagger\) & Res101 & 79.1 \\
CFNet \cite{Zhang2019CoOccurrentFI} & Res101 &79.6 \\
AttaNet \cite{Song2021AttaNetAN}& Res101 & 80.5 \\
PSPNet \cite{zhao2017pyramid} & Res101 & 81.2 \\
ANNet \cite{Zhu2019AsymmetricNN} & Res101 & 81.3 \\
PSANet \cite{zhao2018psanet} & Res101 & 81.4 \\
CCNet \cite{huang2019ccnet} & Res101 & 81.4 \\
OCNet \cite{Yuan2021OCNetOC} & Res101 & 81.9 \\
HANet \cite{Choi2020CarsCF} & Res101 & 82.1 \\
RecoNet \cite{Chen2020TensorLR} & Res101 & 82.3 \\
OCR \cite{Yuan2020ObjectContextualRF} & Res101 & \underline{82.4} \\
\textbf{FLANet (Ours)} & Res101 & \textbf{83.0} \\
\hline
\textit{Advanced Backbone}&&\\
SPGNet \cite{Cheng2019SPGNetSP} &\(2\times\)Res50 & 81.1 \\
DANet \cite{fu2019dual} & R101+MG & 81.5 \\
DGCNet \cite{Zhang2019DualGC} & R101+MG & 82.0 \\
ACNet \cite{fu2019adaptive} & R101+MG & 82.3 \\
GFF \cite{Li2020GatedFF} & R101+PPM& 82.3  \\
ACFNet \cite{zhang2019acfnet} & R101+ASPP & 81.8 \\
GALD \cite{Li2019GlobalAT}\(^\ddagger\) &R101+ASPP & 83.3 \\
HRNet \cite{Sun2019DeepHR} & HRNetW48 & 81.6 \\
OCNet \cite{Yuan2021OCNetOC} & HRNetW48 & 82.5 \\
OCR \cite{Yuan2020ObjectContextualRF}\(^\ddagger\) & HRNetW48 & \textbf{84.2} \\
\textbf{FLANet (Ours)} & HRNetW48 & \underline{83.6} \\
\bottomrule[1pt]
\end{tabular}
\caption{Comparison with state-of-the-art models on the Cityscapes test set.  \(\dagger\) and \(\ddagger\) mark the method w/o using Cityscapes val set and the method using Mapillary dataset for training, respectively.}
\label{cityscapes}
\end{table}

To evaluate the proposed FLANet, we conduct extensive experiments on the Cityscapes \cite{cordts2016cityscapes}, the ADE20K \cite{Zhou_2017_CVPR}, and the PASCAL VOC \cite{Everingham2009ThePV}.

\subsection{Datasets}

\subsubsection{Cityscapes} Cityscapes is a dataset for urban scene segmentation, which contains 5K images with fine pixel-level
annotations and 20K images with coarse annotations. The dataset has 19 classes and each image is with \(1024\times 2048\) resolution. The 5K images with fine annotations are further divided into 2975, 500, and 1525 images for training, validation, and testing, respectively. 

\subsubsection{ADE20K} ADE20K is a challenging scene parsing benchmark. The dataset contains 20K/2K images for training and validation which is densely labeled as 150 stuff/object categories.  Images in this dataset are from different scenes with more scale variations.

\subsubsection{PASCAL VOC} PASCAL VOC is a golden benchmark of semantic segmentation, which includes 20 object categories and one background class. The dataset contains 10582, 1449, 1456 images for
training, validation, and testing.

\subsection{Implementation Details}

Our implementation is based on PyTorch, and uses ResNet101 and HRNetW48 pre-trained from ImageNet \cite{russakovsky2015imagenet} as the backbone network. Following prior works \cite{yu2018learning}, we apply the poly learning rate policy where the initial learning rate is multiplied by after each iteration with a momentum coefficient of 0.9. For data augmentation, we apply the random color jittering within the range of [0.5, 1.5], random flipping horizontally, random scaling within the set of [0.5, 0.75, 1.0, 1.25, 1.5, 1.75, 2.0], and cropping to augment the training data. For the results compared with the state-of-the-art methods, we train our model until convergence. For the ablation studies, we set the training epochs as 240 for all models to speed up the training process. Besides, the synchronized batch normalization is used to synchronize the mean and standard deviation of batch normalization across multiple GPUs. For evaluation, the commonly used Mean IoU metric is adopted. 

Particularly, for the Cityscapes dataset, we set the initial learning rate as 1e-2, weight decay as 5e-4, batch size as 8, and crop size as 512 × 1024, respectively. For the results reported on the Cityscapes test set, we utilize the following training steps: we first train our model on train+val set, then we fine-tune our model on coarse data, and finally we keep fine-tuning the model on train+val set to get the final result. For the ADE20K dataset, we set the initial learning rate as 2e-2, weight decay as 1e-4, batch size as 16, and crop size as 520 × 520, respectively. For the PASCAL VOC dataset, we set the initial learning rate as 1e-3, weight decay as 1e-4, batch size as 16, and crop size as 520 × 520, respectively.

\subsection{Experiments on the Cityscapes Dataset}

% \begin{figure}[t]
% \centering
% \includegraphics[width=1\columnwidth]{sota_compare.pdf} 
% \caption{ Qualitative comparisons with the state-of-the-art method on the Cityscapes validation set.}
% \label{fig4}
% \end{figure}

\begin{figure}[t]
\centering
\includegraphics[width=1\columnwidth]{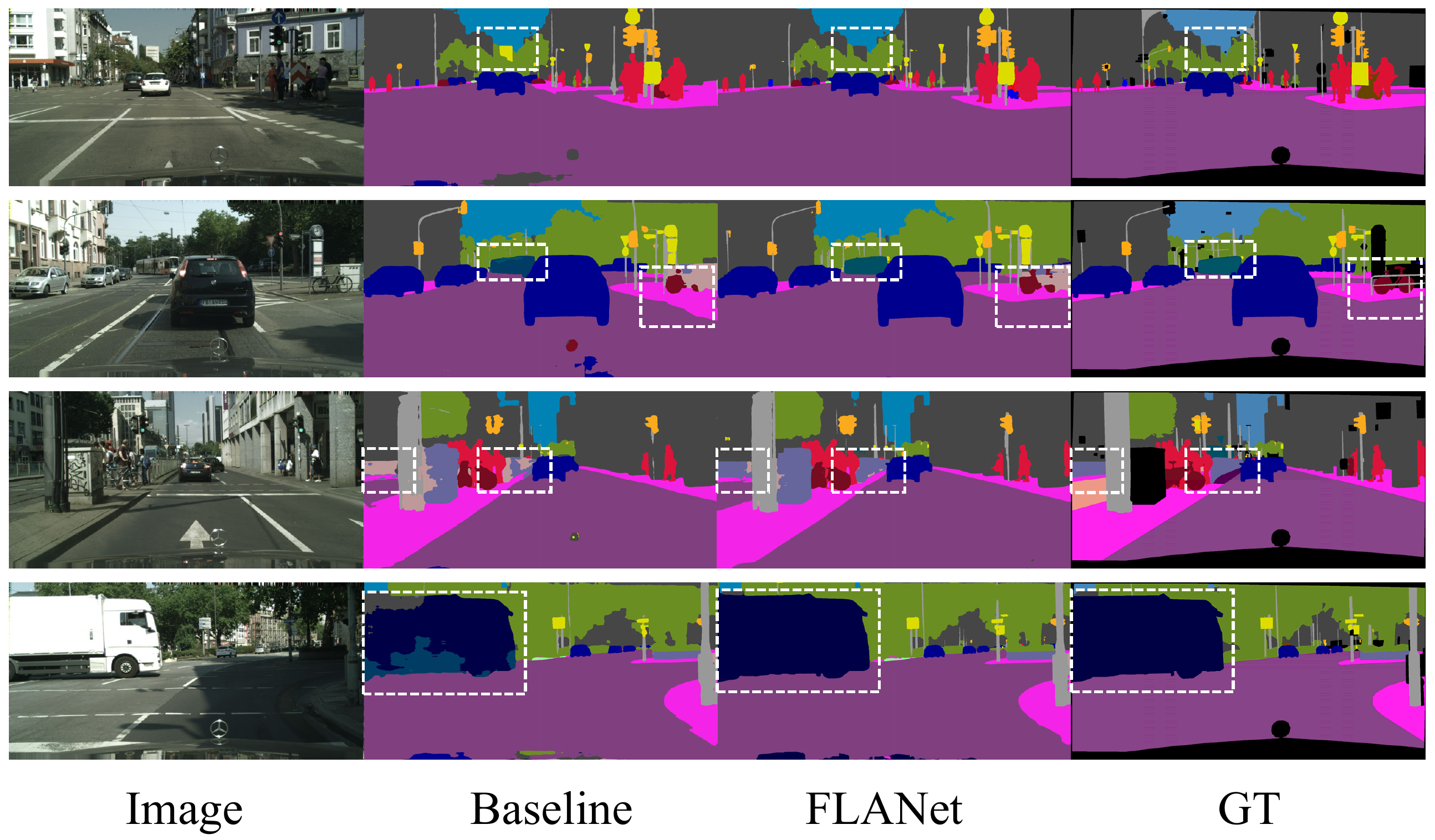} 
\caption{Visualization results of FLANet on the Cityscapes validation set.}
\label{fig4}
\end{figure}

\subsubsection{Comparisons to the State of the Art}
We first compare our proposed method with the state-of-the-art approaches on the Cityscapes test set. And the comparison results are summarized in Tab.\ref{cityscapes}. Among these approaches, the self-attention based models are most related to our method, and more detailed analyses and comparisons will be illustrated in the following subsection. 

From Tab.\ref{cityscapes}, it can be observed that our approach substantially outperforms all the previous techniques based on ResNet101 (Res101) and achieves a new state-of-the-art performance of \textbf{83.0\%} mIoU. Moreover, based on HRNetW48, it achieves a performance that is comparable with methods based on some stronger backbones or even the methods using extra data for training.
% Detailed per-category comparisons are reported in Tab.\ref{per-class}, where our method achieves the highest IoU score on all categories, and large improvements are from categories such as \textit{rider}, \textit{bus}, \textit{train}, and \textit{mbike}. It also proves the benefits of the proposed FLANet in predicting distant objects and maintaining the segmentation consistency inside large objects. 

\begin{table}[t]
\centering
\small
\begin{tabular}{l|cc}
\toprule[2pt]
Method & SS (\%) &MS+F (\%)\\
\midrule[2pt]
ShuffleNetV2& 69.2 &70.8 \\
+FLA& 74.7 &76.3\\
Res18&71.3 &72.5\\
+FLA& 76.5 & 78.1 \\
Res50&74.5 &75.8\\
+FLA& 78.9 &79.7 \\
Res101&75.6 &76.9\\
+FLA&81.9 & 82.6 \\
\bottomrule[1pt]
\end{tabular}
\caption{Ablation study between the baseline and FLANet on Cityscapes validation set according to various backbone networks. \textbf{SS:} Single scale input during evaluation. \textbf{MS:} Multi-scale input. \textbf{F:} Adding left-right flipped input.}\smallskip
\label{ablation}
\end{table}

\begin{figure*}[t]
\centering
\includegraphics[width=1\textwidth]{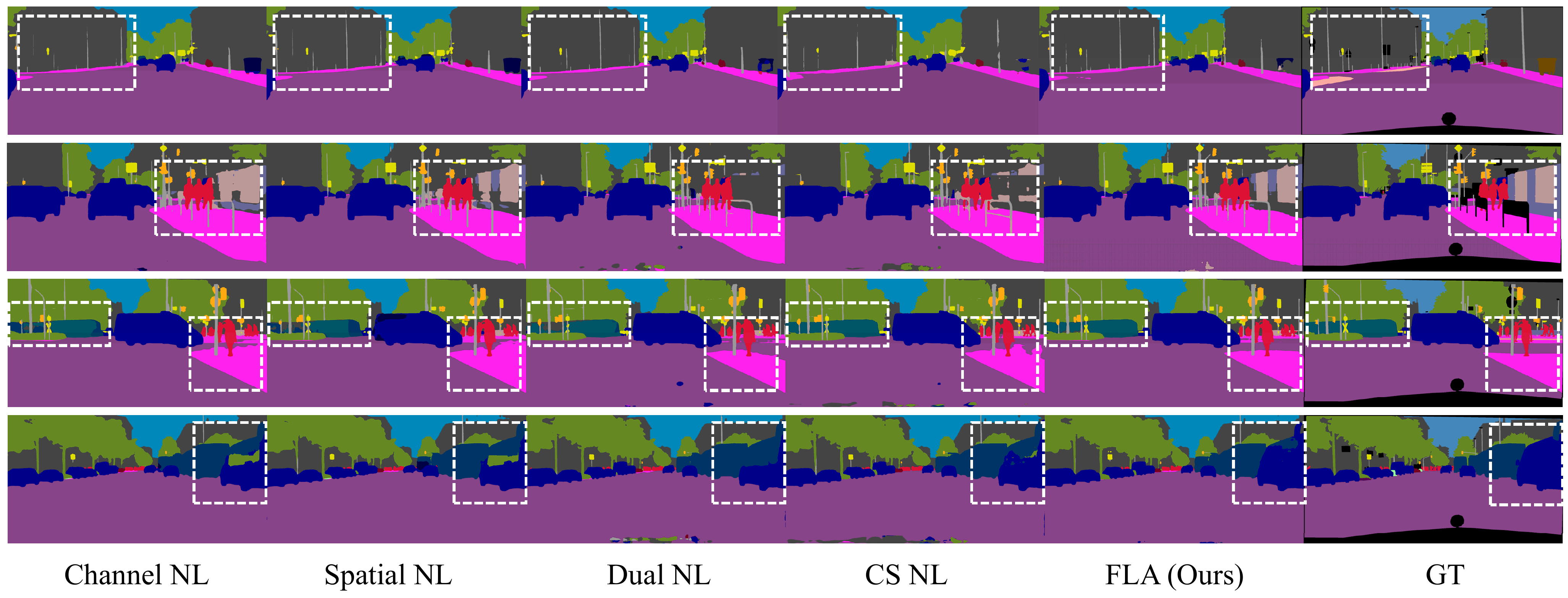} 
\caption{Qualitative comparisons against the NL methods on the Cityscapes validation set. Due to the limited space, we remove the input images and only show the segmentation results and ground truth (GT).}
\label{fig5}
\end{figure*}

\subsubsection{Ablation Studies}
To demonstrate the wide applicability of FLANet, we conduct ablation studies on various backbone networks, including ShuffleNetV2 \cite{ma2018shufflenet} and ResNet series. As listed in Tab.\ref{ablation}, models with FLA consistently outperform baseline models with significant increases no matter what backbone network we use. 

In addition, we provide the qualitative comparisons between FLANet and the Baseline (ResNet50) in Fig.\ref{fig4}, where we use the \textit{white squares} to mark the challenging regions. One can observe that the baseline easily misclassifies those regions but our proposed network is able to correct them. For example, \emph{building} in row 1, \emph{side walk} and distant \emph{train} in row 2, and large \emph{truck} in row 4. It also proves the benefits of the proposed FLANet in predicting distant objects and maintaining the segmentation consistency inside large objects. 

\subsubsection{Comparison with NL Methods} 

\begin{table}[t]
\centering
\small
\resizebox{\columnwidth}{!}{
\begin{tabular}{l|cc|cc}
\toprule[2pt]
Method & SS(\%) &MS+F(\%) &FLOPs(G) &Memory(M) \\
\midrule[2pt]
Baseline &74.5 &75.8 &- &- \\
+EMA & 76.4 & 77.1 &13.91 &98 \\
+Channel NL &  75.6 & 76.9  &\textbf{9.66} &\textbf{40}  \\
\midrule[0.5pt]
+RCCA & 76.7 & 77.8 &16.18 &174  \\
+Spatial NL & 76.3 & 77.6 & 103.90&1320  \\
\midrule[0.5pt]
+Dual NL &  77.1  &  78.0  &113.56 &1378 \\
+CS NL & 77.4  & 78.2 &113.56 &1378 \\
\textbf{+FLA (Ours)}& \textbf{78.9}  & \textbf{79.7} &19.37 & 436 \\
\bottomrule[1pt]
\end{tabular}}
\caption{Detailed comparisons with existing NL models on the Cityscapes validation set. The FLOPs and Memory are computed with the input size 3 × 768 × 768. From up to down, we group the NL methods into three parts, i.e., the channel-only attention, spatial-only attention, and both channel-spatial attention. Adding FLA to the baseline largely increase the mIoU with fewer computation when modeling both channel-wise and spatial-wise relationships.}
\label{nl_compare}
\end{table}

We compare our FLANet with several existing non-local models on the Cityscapes validation set. We measure the increased computation complexity (measured by the number of FLOPs) and GPU memory usage that are introduced by the NL blocks and do not count the complexity from the baselines. To speed up the training procedure, we carry out these comparison experiments on ResNet50, with batch size 8. Besides, we find that, due to the complexity gap, training models with 240 or 480 epochs lead to a great difference in accuracy. For light models like Channel NL, the longer the training time, the worse the testing accuracy. On the contrary, heavy models perform better when using larger training steps. Therefore, to eliminate the influence of training epochs on models with different complexity, we train these NL models two times with 240 or 480 epochs respectively, and then average two scores as the final mIoU results.  

Specifically, the NL models compared in Tab.\ref{nl_compare} include 1) Expectation-Maximization Attention in EMANet \cite{li2019expectation}, donated as “+EMA”; 2) Recurrent Criss-Cross Attention (R=2) in CCNet \cite{huang2019ccnet}, donated as “+RCCA”; 3) two typical NL blocks introduced in Sec.1., donated as “+Channel NL” and “+Spatial NL” respectively; 4) two connection modes introduced in Sec.1., donated as “+Dual NL” and “+CS NL” respectively. Besides, according to whether calculate the channel-only attention, spatial-only attention, and both channel-spatial attention, Tab.4 is divided into three groups.

As illustrated in Tab.\ref{nl_compare}, FLA outperforms these NL methods by a large margin, and the complexity comparison results indicate that the cost of adding FLA is practically negligible even compared with the lightweight-designed models like EMA and RCCA. Moreover, it can be found that the increased computational cost of FLA for capturing spatial attentions is \textbf{the lowest} (about 9.71 GFLOPs) compared with all these spatial-modeling NLs. Even when compared with the Channel NL who requires the lowest computational cost, our FLA outperforms it by \textbf{2.8\%} with the minimum computational increment. And the computational complexity of FLA is consistent to our previous theoretical analysis in Sec.3.3. It is noted that our FLA significantly reduces GFLOPs by about \textbf{83\%} and only requires \textbf{34\%} GPU memory usage of DANet (Dual NL) and CS NL when modeling both channel-wise and spatial-wise relationships. Therefore, FLA has a great advantage of not only an effective way of improving segmentation accuracy but also a lightweight algorithm design for practical usage.

\subsubsection{The Efficacy of FLA}
To further prove that our method can successfully solve the attention missing issue, we also present several qualitative comparison results in Fig.\ref{fig5}. As shown in Fig.\ref{fig5}, we can find that Dual NL and CS NL can combine the advantages of Channel NL and Spatial NL to some extent and generates better segmentation results. However, it is obvious that sometimes they obtain wrong predictions even when they are correctly classified in Channel NL and Spatial NL, such as the examples shown in the second row. This coincides with our claims that the attention missing issue would distort interactions between dimensions and can not be solved by stacking different NL blocks. Compared with those NL methods, accuracies of predictions for both distant/thin categories (e.g., \textit{poles} in the first row) and the ability to maintain the segmentation consistency inside large objects (e.g., \textit{train} in the third row and \textit{car} in the last row) are significantly improved after using the proposed FLA. And the quantitative per-class comparisons can be seen in Fig.\ref{fig2}. This phenomenon can also demonstrate that FLA can optimally model both channel-wise and spatial-wise relations only by using a single non-local block.

\subsubsection{Visualization of Attention Module}

\begin{figure}[t]
\centering
\includegraphics[width=1\columnwidth]{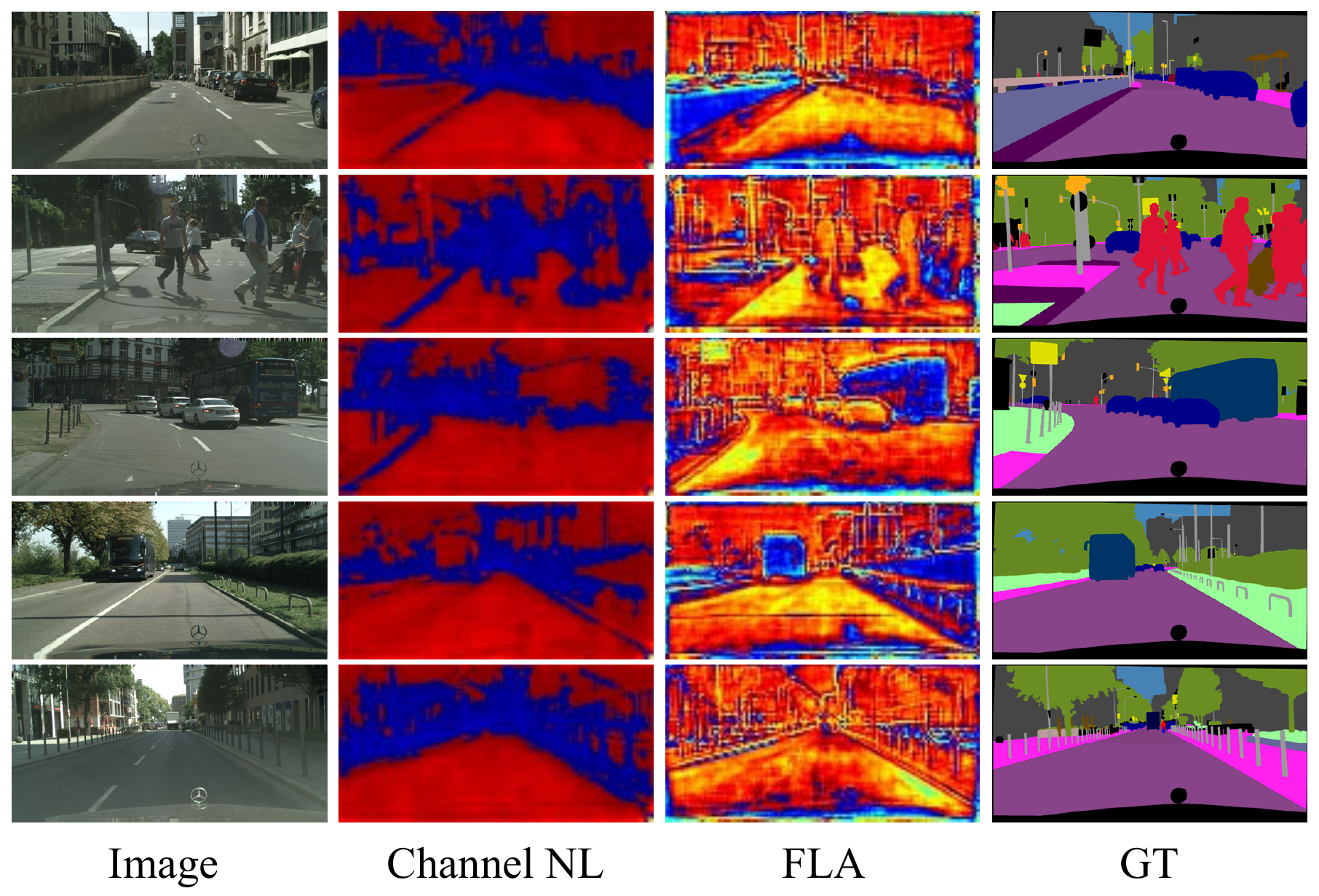} 
\caption{Visualization of attended feature maps of Channel NL and our FLA on the Cityscapes validation set, where feature maps are visualized by averaging along the channel dimension.}
\label{fig6}
\end{figure}

To get a deeper understanding of the effectiveness of our FLA in encoding spatial attentions into the channel affinity map, we visualize the attended feature maps and analyze how FLA improves the final result. We also visualized the attended feature maps of the traditional Channel NL for further comparison. As shown in Fig.\ref{fig6}, both Channel NL and our FLA highlight some semantic areas and guarantee consistent representation inside large objects like roads and buildings. Furthermore, it is noted that the attended feature maps of FLA are more structured and detailed than that of Channel NL. For example, distant poles and object boundaries are highlighted for all images. Particularly, FLA can also distinguish different classes, e.g., bus and car in the third row. These visualization results further demonstrate that our proposed module can capture and encode the spatial similarities into the channel attention map to achieve full attentions. 

\begin{table}[t]
\centering
\small
\begin{tabular}{lcc}
\toprule[2pt]
Method & Backbone & mIoU\\
\midrule[2pt]
\textit{Simple Backbone}&&\\
AttaNet \cite{Song2021AttaNetAN}& Res101 & 43.71 \\
CCNet \cite{huang2019ccnet} & Res101   & 45.22\\
ANNet \cite{Zhu2019AsymmetricNN} & Res101   &  45.24\\
OCR \cite{Yuan2020ObjectContextualRF} & Res101 & 45.28 \\
GFF \cite{Li2020GatedFF}&Res101&  45.33\\
OCNet \cite{Yuan2021OCNetOC} & Res101   &  45.40\\
DMNet \cite{He2019DynamicMF} & Res101 &    45.50\\
RecoNet \cite{Chen2020TensorLR}& Res101  &45.54\\
DNL \cite{Yin2020DisentangledNN} & Res101 &   45.97\\
CPNet \cite{Yu2020ContextPF} & Res101 & \underline{46.27}\\
\textbf{FLANet (Ours)} &Res101 &\textbf{46.68}\\
\hline
\textit{Advanced Backbone}&&\\
DANet \cite{fu2019dual} & Res101+MG   &  45.22\\
ACNet \cite{fu2019adaptive}  & Res101+MG &   \underline{45.90} \\
HRNetV2  \cite{Sun2019DeepHR} & HRNetW48  &  42.99\\
OCNet \cite{Yuan2021OCNetOC} & HRNetW48   &  45.50\\
OCR \cite{Yuan2020ObjectContextualRF} & HRNetW48 & 45.66 \\
DNL  \cite{Yin2020DisentangledNN} & HRNetW48& 45.82\\
\textbf{FLANet (Ours)} &HRNetW48 &\textbf{46.99}\\
\bottomrule[1pt]
\end{tabular}
\caption{Comparisons on the ADE20K validation set.}
\label{ade}
\end{table}

\begin{table}[t]
\centering
\small
\begin{tabular}{lcc}
\toprule[2pt]
Method & Backbone & mIoU\\
\midrule[2pt]
\textit{Simple Backbone}&&\\
DeepLabv3 \cite{chen2017rethinking} & Res101   &85.7 \\
EncNet \cite{Zhang2018ContextEF} & Res101   & 85.9 \\
DFN \cite{yu2018learning} & Res101   & 86.2\\
CFNet \cite{Zhang2019CoOccurrentFI}&Res101& 87.2 \\
EMANet \cite{li2019expectation} &Res101&87.7\\
RecoNet \cite{Chen2020TensorLR}& Res101  &\underline{88.5}\\
DeeplabV3+ \cite{chen2018encoder}  & Xception &\textbf{89.0}  \\
\textbf{FLANet (Ours)} &Res101 &87.9\\
\hline
\textit{Advanced Backbone}&&\\
EMANet \cite{li2019expectation} &Res150&88.2\\
RecoNet \cite{Chen2020TensorLR}& Res150  &\textbf{89.0}\\
\textbf{FLANet (Ours)} &Res150 &\underline{88.5}\\
\bottomrule[1pt]
\end{tabular}
\caption{Comparisons on the PASCAL VOC dataset.}
\label{pascal_voc}
\end{table}

\subsection{Experiments on the ADE20K Dataset}

To further validate the effectiveness of our FLANet, we conduct experiments on the ADE20K dataset, which is a challenging scene parsing dataset with both indoor and outdoor images. Tab.\ref{ade} reports the performance comparisons between FLANet and the state-of-the-art models on the ADE20K validation set. Our approach achieves 46.99\% mIoU score, outperforms the previous state-of-the-art methods by 0.72\%, which is significant due to the fact that this benchmark is very competitive. CPNet achieves previous best performance among those methods and utilizes the learned context prior with the supervision of the affinity loss to capture the intra-class and inter-class contextual dependencies. In contrast, our FLANet try to capture both spatial-wise and channel-wise dependencies in a single attention map and achieve better performance.

\subsection{Experiments on the PASCAL VOC Dataset}

To verify the generalization of our proposed FLANet, we conduct experiments on the PASCAL VOC dataset. The comparison results are shown in Tab.\ref{pascal_voc}. FLANet based on ResNet101 and ResNet150 achieves comparable performance on the PASCAL VOC test set.

\section{Conclusions and Future Work}
In this paper, we find that traditional self-attention methods suffer from the attention missing problem caused by matrix multiplication. To mitigate this issue, we reformulate the self-attention mechanism into a fully attentional manner, which can capture both channel and spatial attentions with a single attention map and also with much less computational complexity. Specifically, we construct global contexts to introduce spatial interactions into the channel attention maps. Our FLANet achieves outstanding performance on three semantic segmentation datasets. Besides, we also consider the way of introducing channel interactions into the traditional Spatial NL. However, the extremely high computational load limits its practical application. In the future, we will try to achieve that in a more efficient way.

\section{Acknowledgments}
This work was supported in part by Shenzhen Natural Science Foundation under Grant JCYJ20190813170601651, and in part by Shenzhen Institute of Artificial Intelligence and Robotics for Society under Grant AC01202101006 and Grant AC01202101010.

\bibliographystyle{aaai22}
\bibliography{ref}

\end{document}